\documentclass[10pt,twocolumn,letterpaper]{article}

\usepackage{cvpr}              

\usepackage[dvipsnames]{xcolor}

\definecolor{cvprblue}{rgb}{0.21,0.49,0.74}
\usepackage[pagebackref,breaklinks,colorlinks,citecolor=cvprblue]{hyperref}

\def\paperID{7143} 
\def\confName{CVPR}
\def\confYear{2024}

\title{
\DriveLVLM{}: A General World Model for Autonomous Driving
}

\author{
Fan Jia$^{1\star}$ \hspace{0.45cm} 
Weixin Mao$^{2\star\dagger}$ \hspace{0.45cm} Yingfei Liu$^{1}$ \hspace{0.45cm} Yucheng Zhao$^{1}$ \hspace{0.45cm} 
Yuqing Wen$^{3\dagger}$ \\
Chi Zhang$^{4}$ \hspace{0.45cm} Xiangyu Zhang$^{1}$ \hspace{0.45cm} Tiancai Wang$^{1\ddagger}$ \\
[.5ex] $^1$MEGVII Technology \hspace{0.9cm} $^2$Waseda University  \hspace{0.9cm} \\
$^3$University of Science and Technology of China \hspace{0.9cm} $^4$Mach Drive \\ [.5ex]
}

\usepackage{multirow}
\usepackage{colortbl}

\usepackage[most]{tcolorbox}  
\usepackage{lipsum}       
\newtcolorbox{myquote}{colback=orange!10, colframe=blue!100, fontupper=\itshape, arc=4mm, boxrule=0.5pt}

\newcommand\blfootnote[1]{%
  \begingroup
  \renewcommand\thefootnote{}\footnote{#1}%
  \addtocounter{footnote}{-1}%
  \endgroup
}

\newcommand{\DriveLVLM}{ADriver-I}

\usepackage{pifont}
\newcommand{\cmark}{\ding{51}}%
\newcommand{\xmark}{\ding{55}}%

\usepackage{epigraph}

\definecolor{darkgreen}{rgb}{0.52, 0.64, 0.43} 
\definecolor{blue1}{rgb}{0.27, 0.56, 0.77}
\definecolor{orange1}{rgb}{0.93, 0.49, 0.19}

\begin{document}
\maketitle
\begin{abstract}
Typically, autonomous driving adopts a modular design, which divides the full stack into perception, prediction, planning and control parts. Though interpretable, such modular design tends to introduce a substantial amount of redundancy. Recently, multimodal large language models (MLLM) and diffusion techniques have demonstrated their superior performance on comprehension and generation ability. In this paper, we first introduce the concept of interleaved vision-action pair, which unifies the format of visual features and control signals. Based on the vision-action pairs, we construct a general world model based on MLLM and diffusion model for autonomous driving, termed \DriveLVLM{}. It takes the vision-action pairs as inputs and autoregressively predicts the control signal of current frame. The generated control signals together with the historical vision-action pairs are further conditioned to predict the future frames. With the predicted next frame, \DriveLVLM{} performs further control signal prediction. Such process can be repeated for infinite times, \DriveLVLM{} achieves the autonomous driving in the world created by itself. Extensive experiments are conducted on nuScenes and our large-scale private datasets. \DriveLVLM{} shows impressive performance compared to several constructed baselines. We hope our \DriveLVLM{} can provide some new insights for future autonomous driving and embodied intelligence.
\end{abstract}

\blfootnote{
$^{\star}$ Equal contribution. $\dagger$ During the internship at MEGVII Technology. $^{\ddagger}$ Corresponding author.
}

\section{Introduction}
\label{sec:intro}
\epigraph{The future depends on what we do in the present.}{Mahatma Gandhi}

Recently, autonomous driving has shown great progress thanks to the large development on BEV perception~\cite{liu2022petr,liu2022petrv2,wang2022detr3d,wang2023streampetr} and end-to-end planning~\cite{prakash2021multi, wu2022trajectory, hu2022st, hu2023uniad, jiang2023vad, chitta2021neat}. Typically, the autonomous driving system can be divided into four components: perception, prediction, planning and control. The perception module is usually used to detect or track the surrounding vehicles, pedestrian or lanes while the prediction module is employed to predict the future trajectories of dynamic objects. Based on the perception and prediction results, the planner further predicts the planning positions of ego car like waypoints and control system generates the corresponding low-level control signals. Such sequential system, which adopts a modular design, sounds reasonable since each module is interpretable. It is relatively easy to trace back to the reason when accidents occur.

Compared to the mainstream autonomous driving system, human drivers mainly share two key differences. The first one is that human drivers tend to directly take actions based on the captured visual information, without rely on the sequential pipeline above. They simply adjust the steer and pedal
in an end-to-end manner without introducing too much consideration. Moreover, human drivers are able to predict the near future. For example, when we see the traffic light signal changing from green to yellow, we predict that traffic light is about to turn red and we need to slow down and prepare to stop. So we wonder if it is possible to construct a system that directly outputs control signals and predicts the future scene, similar to human driver.

\begin{table*}[!h]
    \centering
    \begin{tabular}{lcc|cc}
        \toprule
        \multirow{2}{*}{Method} & \multicolumn{2}{c}{Input prior} & \multicolumn{2}{c}{Output} \\ 
  \cmidrule(l){2-3} \cmidrule(l){4-5}  
        & w/o Box & w/o HD Map & Control Signal & Predict Future \\
         \midrule
         DriveGPT4~\cite{DriveGPT4} & \xmark & \cmark & \cmark &  \xmark \\
         GAIA-1~\cite{hu2023GAIA} & \cmark & \cmark & \xmark & \cmark \\
         DriveDreamer~\cite{wang2023drivedreamer} & \xmark & \xmark & \cmark & \cmark \\ 
        \rowcolor[gray]{.9}  ADriver-I & \cmark & \cmark & \cmark & \cmark \\ 
         \hline
    \end{tabular}
    \caption{Overall comparison with existing methods. Different from existing approaches, ADriver-I directly outputs the control signal of current frame. It further predicts the future frames based on the predicted control signal without any bounding box or HD map prior.}
    \label{tab:ref_frame} 
\end{table*}


Multimodal large language models (MLLMs)~\cite{llava, llava1.5, zhu2023minigpt, gpt4v} and large language models (LLMs) ~\cite{touvron2023llama, touvron2023llama2, du2022glm, zeng2022glm, 2018GPT1, 2019GPT2, 2019GPT3, gpt4} have attracted attention due to their excellent performance in logical reasoning and generalization capabilities. However, they mainly focus on dialogue with images with different finegrain levels. In this paper, we build a novel system called ADriver-\uppercase\expandafter{\romannumeral1} to unify control signal prediction and future scene generation. Inspired by the interleaved document in MLLMs, we introduce the interleaved vision-action pair to unify the format of visual features and corresponding control signals. The control signals, such as the steer angles and ego speed, can be converted into text-like expressions. 
\DriveLVLM{} takes the historical vision-action pairs and current visual tokens as inputs and directly predicts the control signals of current frame. Conditioned on historical vision-action pairs and the predicted actions, the video diffusion model (VDM) is further employed to predict the scene of following frames. It means the predicted current action directly affects the scene in the future.

With the unified action prediction and future generation framework, ADriver-I can perform the autonomous driving under the infinite scenario generated by itself, like living in its own world. Considering the strong generality of MLLMs, \DriveLVLM{} system can be easily generalized to the real world. ADriver-I
shows impressive performance on nuScenes and our private datasets. Given three historical vision-action pairs, the L1 error including speed and steer angle of current frame is only 0.072 m/s and 0.091 rad. The FID and FVD metrics for generated future four frames are 5.52 and 97 respectively.

\section{Related Works}
\label{sec:relatedWorks}


\subsection{Multimodal Large Language Models}

Recently, large language models (LLMs) and multimodal large language models (MLLMs) have attracted a lot of attention due to the powerful comprehension and generation capabilities. 
MLLMs are built on top of LLMs like LLaMA~\cite{touvron2023llama, touvron2023llama2}, Vicuna~\cite{peng2023vicuna, vicuna1.5} and GPT~\cite{2018GPT1, 2019GPT2, 2019GPT3}. These models can receive different modal inputs such as text, image, and video. Some typical methods such as LLaVA~\cite{llava, llava1.5}, miniGPT4~\cite{zhu2023minigpt} and BLIP-2~\cite{23blip2}, use the image and text tokens as input and are trained to achieve the cross-modal understanding. 
MLLMs can be extended for video-text~\cite{luo2023valley, li2023videochat, wang2023chatvideo, zhang2023videollama} and audio-text~\cite{rubenstein2023audiopalm, le2023voicebox} comprehension.

The development of MLLMs greatly inspires the research on embodied intelligence~\cite{huang2023voxposer, jiang2022vima, PaLM-E, RT-1, RT-2} and autonomous driving. For example, VIMA~\cite{jiang2022vima} utilizes the Mask RCNN~\cite{2017maskrcnn} to extract object regions. Those object regions together with the text description are input to transformer architecture to  predict the motor actions. 
VoxPoser~\cite{huang2023voxposer} uses pretrained MLLM and LLM to generate a value map for motion planning without needing extra training. Combining pretrained PaLM~\cite{chowdhery2022palm} and ViT-22B~\cite{2023vit22b}, PaLM-E can perform multiple tasks, such as motion planning, tabletop manipulation and image description. 
Furthermore, RT-2~\cite{RT-2} proposes vision-language-action (VLA) models for robot control and directly outputs low-level control signals.
As our concurrent work, DriveGPT4~\cite{DriveGPT4}, based on MLLM, takes the video and text as input. It can output the control signals and provide the corresponding reason for interpretability. 



\subsection{End-to-end Autonomous Driving}
Typically, autonomous driving (AD) can be divided into perception, prediction, planning and control. 
Most state-of-the-art end-to-end autonomous methods~\cite{prakash2021multi, wu2022trajectory, hu2022st, hu2023uniad, jiang2023vad, chitta2021neat} adopt the encoder-decoder paradigm to extract information from raw sensor data and predict the planning results. Transfuser~\cite{prakash2021multi} and TCP~\cite{wu2022trajectory} directly predict planning results (way points) without constructing any scene representation.
Other methods use various scene representations to assist the model in understanding the driving rules in real world.
ST-P3~\cite{hu2022st} builds a dense cost map from the semantic map and then uses the hand-crafted rules to get the best planning trajectory with minimum cost. UniAD~\cite{hu2023uniad} integrates diverse scene representations in a hierarchical manner including the segmentation map, motion flow map and BEV occupancy map. 
VAD~\cite{jiang2023vad} adopts a fully vectorized approach that employs the vectorized agent motion and map, eliminating the need for computationally intense rasterized representations.

\begin{figure*}[t]
  \centering
   \includegraphics[width=1.0\linewidth]{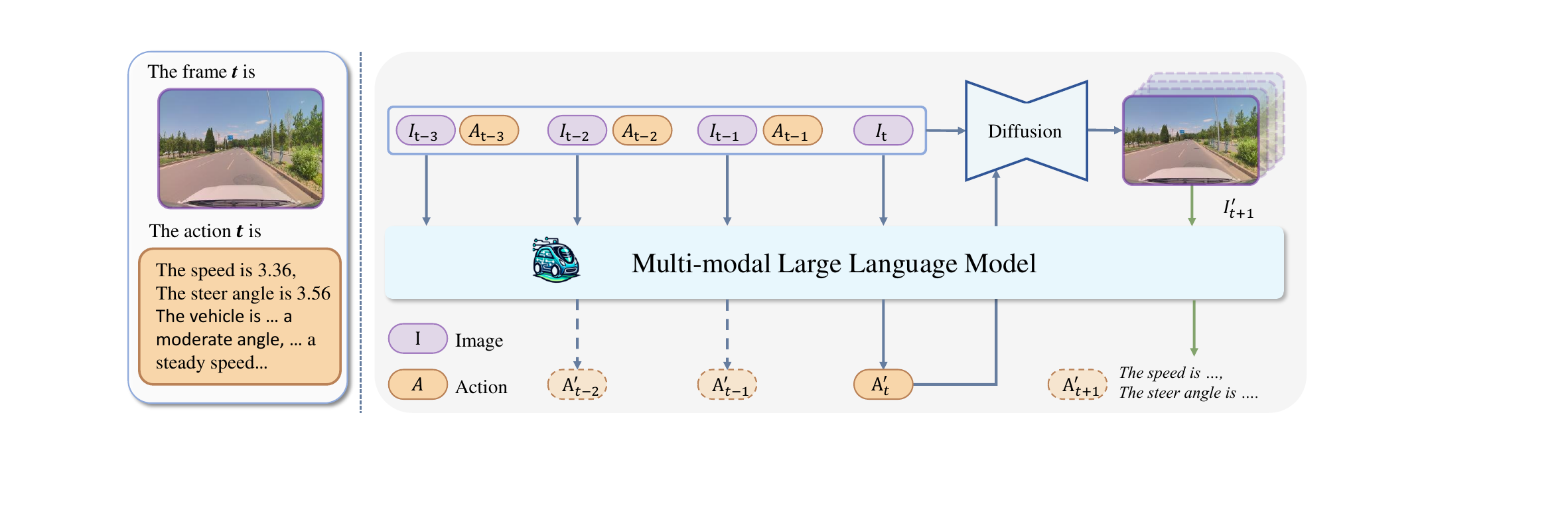}
   \caption{Overview of of our \DriveLVLM{} framework. It takes the historical interleaved vision-action pairs \{$I$, $A$\} and current visual token as inputs. The multi-modal large language model (MLLM) reasons out the control signal $A_{t}$ of current frame. The predicted action $A_{t}$ is further used as the condition prior of video latent diffusion model (VDM) to generate the future four frames. The predicted next frame $I^{'}_{t+1}$ is selected and further input to the MLLM to produce the control signal $A^{'}_{t+1}$. Such process (\textcolor{darkgreen}{green line}) can be repeated for infinite times and it achieves the self-learning in the world created by itself. The dashed lines represent the output only in training process.}
   \label{fig:arch}
\end{figure*}

\subsection{Generative Models for Autonomous Driving}


In the field of autonomous driving, scene generation advances through various methods, including generative adversarial networks (GANs) ~\cite{goodfellow2014GAN}, variational auto-encoders (VAEs)~\cite{2014VAE, van2017VQVAE} and diffusion models~\cite{20DDPM, 21DDIM}.
DriveGAN~\cite{21DriveGAN} predicts future driving videos by correlating driving actions with pixel changes. Similarly, BEVGen~\cite{23bevgen}, built upon VQ-VAE~\cite{van2017VQVAE}, adeptly creates multi-view images from Bird's Eye View (BEV) layouts. BEVControl~\cite{yang2023bevcontrol} goes a step further, generating both foreground and background in street View images, even accommodating sketch style inputs.
Building on the Stable Diffusion~\cite{Rombach_2022_CVPR}, video latent diffusion model~\cite{blattmann2023align} is proposed for synthesizing high-resolution video, exhibiting its astonishing generation qualities. 
Furthering this progress, recent methodologies for controllable driving-scene video generation have been developed.
Panacea~\cite{Anonymous} proposes a layout-conditioned video generation system aimed at diversifying the data sources for training perception models. 

\subsection{World Models for Autonomous Driving}
In terms of world model, there are two kinds of definitions: one is to purely predict the future and another one is unifying the action prediction and future generation. 
In the fields of reinforcement learning and robotics~\cite{22DayDreamer, 20GameGAN, seo2023masked, xie2023citydreamer, HaS18Recurrentworld, 20DreamtoControl}, world models are often used to predict how the environment will respond to the actions of agents. These models can be based on a variety of data (e.g., RGB images, depth images, point clouds, etc.) to understand the environment's behavior and predict future states. 



GAIA-1~\cite{hu2023GAIA} proposes a generative world model that takes video, text, and action as inputs to generate realistic driving scenarios. 
DriveDreamer~\cite{wang2023drivedreamer} also introduce a world model by generating the future driving scenarios and predicting the control signals. 
Our work is closely related to GAIA-1 and DriveDreamer. However, there are some major differences. GAIA-1 is more like a generator for driving scenarios while ignoring the control signal prediction. DriveDreamer relies heavily on rich prior information, like high-definition (HD) maps and 3D bounding boxes, for future generation. In contrast, our \DriveLVLM{} unifies the control signal prediction and the future scene generation. For future generation, it eliminates the need for extensive prior information. To the best of our knowledge, we are the first that introduces the concept of infinite driving. \DriveLVLM{} achieves the infinite driving in the world created by itself. 



\section{Method}
\label{sec:method}
In this section, we first describe the overall architecture of \DriveLVLM{} in Sec.~\ref{sec:arch}. Then we introduce the prompt construction details in Sec.~\ref{sec:prompt}. Finally, the model training details are provided in Sec.~\ref{sec:training}.

\begin{figure*}[t]
  \centering
   \includegraphics[width=1.0\linewidth]{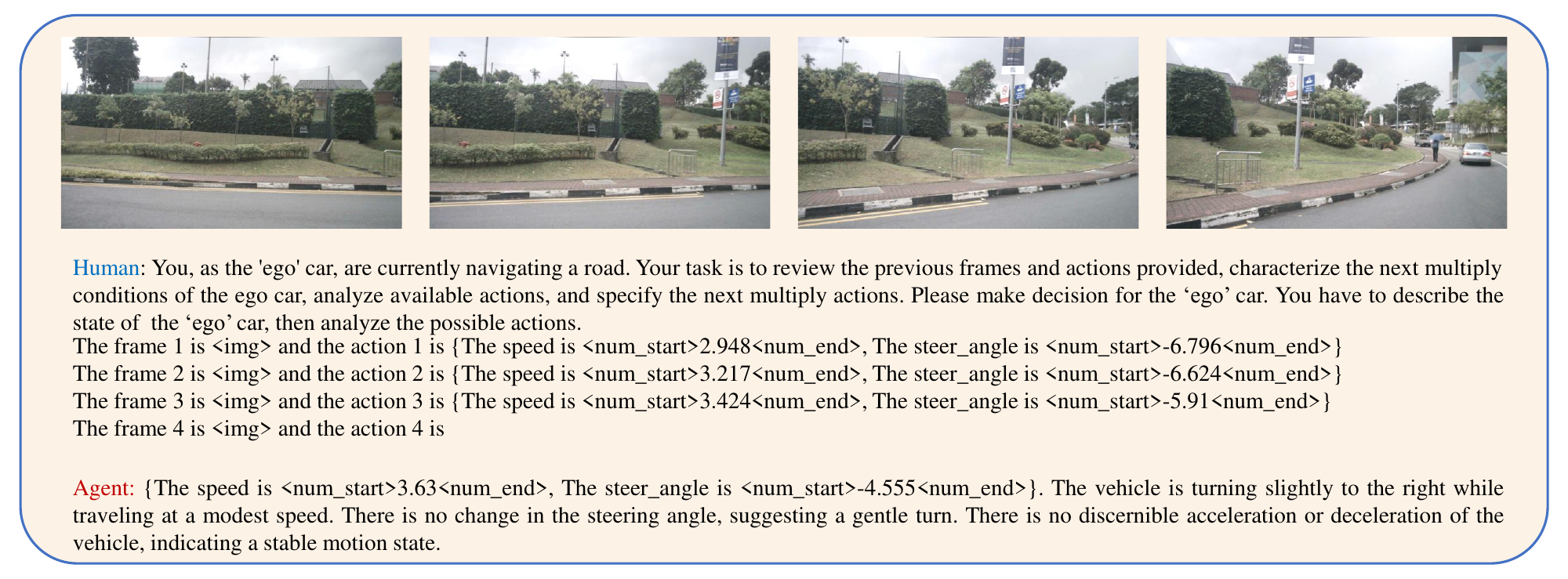}
   \vspace{-15pt}
   \caption{An example of conversation used for multimodal large language model. The human prompt mainly contains system prompt and the interleaved vision-action pair. The agent answer is the control signal including the speed and steer angle.$\mathtt{<img>}$ is the visual token. $\mathtt{<num\_start>}$ and $\mathtt{<num\_end>}$ are the beginning and ending tokens for the digits, respectively.}
   \label{fig:conversation}
\end{figure*}

\subsection{Architecture} 
\label{sec:arch}

We introduce \DriveLVLM{}, a general world model for autonomous driving based on a multimodal large language model (MLLM) and video diffusion model (VDM). The overall framework of \DriveLVLM{} is shown in Fig.~\ref{fig:arch}.

\noindent \textbf{Overall Pipeline:}
The current video frame $I_t$ and the historical visual-action pairs $(I_{t-1}, A_{t-1}), ..., (I_{t-3}, A_{t-3})$ serve as the inputs for both MLLM and VDM.
During training process, MLLM outputs the low-level control signal set \{$A^{'}_{t-2}$, $A^{'}_{t-1}$, $A^{'}_t$\} in an auto-regressive manner. They are all supervised by the corresponding control signals. 
After that, the output control signal $A^{'}_t$ is further used as the prompt of VDM to predict the next four frames $\{I^{'}_{t+1}, ..., I^{'}_{t+4}\}$. The predicted frames are supervised by the ground-truths. During inference, MLLM directly outputs the control signal of current frame $A^{'}_t$ with a single step. VDM follows the same process and predicts the future frames.
One important thing for ADriver-I is that it provides an attractive possibility for future autonomous driving. The generated next frame $I^{'}_{t+1}$ is served as the "current frame" in the next timestamp and further input to MLLM to produce the $A^{'}_{t+1}$. The steps above can be repeated cyclely. In this way, it achieves the infinite driving in the world generated by itself. 

\noindent \textbf{Multi-modal Large Language Model:} MLLM consists of three modules: a pre-trained large language model (LLM), an visual encoder and an visual adapter. We adapt Vicuna-7B-1.5~\cite{vicuna1.5} as the LLM. Vicuna is fine-tuned on LLAMA2~\cite{touvron2023llama2}. 
We utilize CLIP-ViT-Large~\cite{CLIP} as the visual encoder, pretrained on a huge amount of image-text pairs. Two multi-layer perceptron (MLP) layers are employed as the visual adapter, pretrained by LLaVA-7B-1.5~\cite{llava1.5}, to align the visual features with language features.

\noindent \textbf{Video Diffusion Model:}
We construct our VDM based on a latent diffusion model~\cite{rombach2022high} for video generation. It is built upon Stable Diffusion 2.1~\cite{Rombach_2022_CVPR} and enhanced with temporal-awareness modules similar to those in video latent diffusion model~\cite{blattmann2023align}. We enrich the model with history-conditioned functionality by integrating a reference-video control, which concatenates given frames with the diffusion input. Additionally, text condition modules are retained to enable action-guided future generation. In summary, our VDM integrates control signals with historical frames, serving as an active generator for our MLLM.

\begin{figure}[t]
  \centering
   \includegraphics[width=1.0\linewidth]{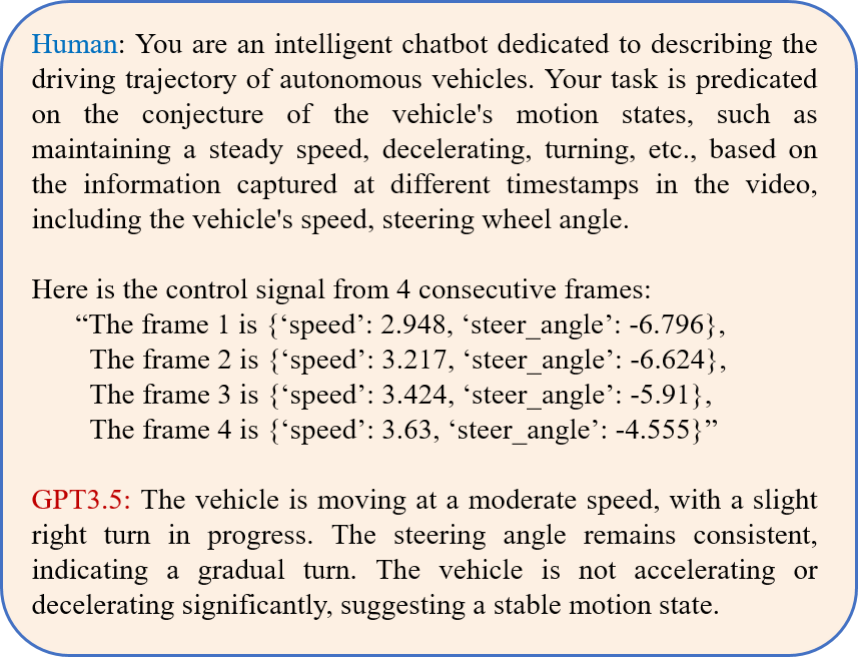}
   \caption{An example of conversation used for guiding GPT3.5 to generate the corresponding motion prompt. The control signals of the historical frame and current frame are used as inputs, while GPT3.5 outputs potential motion states.}
   \label{fig:gpt}
\end{figure}

\subsection{Prompt Construction}
\label{sec:prompt}

we obtain the front-view video frames and their corresponding low-level control signals (e.g., speed and steering angle) from our private dataset and nuScenes dataset.

\begin{figure*}[t]
  \centering
   \includegraphics[width=1.0\linewidth]{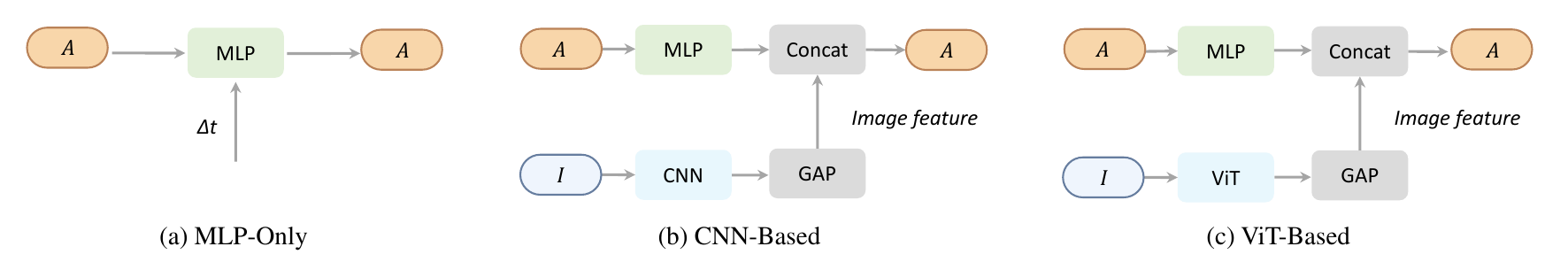}
   \vspace{-15pt}
   \caption{Overall architectures of some compared baselines. The baseline (a) takes the historical action sequence as input and uses a MLP network to predict the action of current frame. Based on (a), the baseline (b) extracts the image features by a CNN and global average pooling (GAP), after which the image features are concatenated with the action features to produce the current action. For the baseline (c), the CNN in (b) is replaced by a ViT-B backbone.}
   \label{fig:multi-baseline}
\end{figure*}

\begin{table*}[t]
   
  \renewcommand\tabcolsep{4.6pt} 
  \small
  \label{control_compare}
  \centering
\begin{tabular}{lcccccccccc}
\toprule
    \multirow{2}{*}{Method} & \multicolumn{5}{c}{Speed (m/s)}            & \multicolumn{5}{c}{Steer angle (rad)}      \\ \cmidrule(l){2-6} \cmidrule(l){7-11}  
 & L1$\downarrow$  & $A_{0.01}\uparrow$  & $A_{0.03}\uparrow$ & $A_{0.05}\uparrow$ & $A_{0.07}\uparrow$  & L1$\downarrow$  & $A_{0.01}\uparrow$  & $A_{0.03}\uparrow$ & $A_{0.05}\uparrow$ & $A_{0.07}\uparrow$\\ 
    \midrule
    MLP-Only & 0.122 & 0.189 & 0.275 & 0.361 & 0.440 & 0.101 & 0.183 & 0.482 & 0.641 & 0.715 \\
    CNN-Based & 0.106 & 0.191 & 0.301 & 0.407 & 0.474  & 0.095 & 0.277 & 0.527 & 0.648 &  0.721\\
    ViT-Based & 0.103 & 0.200 & 0.326 & 0.438 & 0.489 & 0.092 & 0.299 & 0.546 & 0.656 & 0.724\\
    \midrule
    \rowcolor[gray]{.9} \DriveLVLM{} & 0.072 & 0.237 & 0.398 & 0.535 & 0.640 &  0.091 & 0.411 & 0.575 & 0.664 & 0.731 \\
    \rowcolor[gray]{0.9} \DriveLVLM{}$\dag$  & 0.035 & 0.295 & 0.519 & 0.790 & 0.862 & 0.015 & 0.643 & 0.840 & 0.925 & 0.964  \\
    \bottomrule
  \end{tabular}
  \vspace{-5pt}
  \caption{Quantitative comparison is conducted from two perspectives: \emph{cross-model} and \emph{cross-dataset}. For cross-model comparison, we compare \DriveLVLM{} with three baselines on the nuScenes dataset. For cross-dataset comparison, we compare the performance of \DriveLVLM{} on nuScenes and our private datasets. $\dag$ represents the results on our private dataset.
  }
  \label{tab:quantitative}
\end{table*}

\noindent \textbf{Prompt for MLLM.} 
we convert the low-level control signal into text so that it can be processed by LLM as language. The action tokens $\mathtt{<TOKEN_{act}>}$ are further obtained by language tokenizer. 
Each video frame is featured by the CLIP-ViT-Large~\cite{CLIP} and further processed by the visual adapter to produce the visual tokens $\mathtt{<TOKEN_{img}>}$.
Visual tokens of each video frame paired with the corresponding action tokens forms the \emph{interleaved vision-action pair}. The introduced vision-action pair has the following advantages. (1) It enables the  multi-round conversation to adapt unfixed frame length, improving the flexibility of MLLM. (2) It unifies the interleaved future generation and action prediction under the word embedding space.
As illustrated in Fig~\ref{fig:conversation}, despite of the visual and action tokens, we also introduce the system prompt to describe the background, guiding the reasoning mode of MLLM.
Overall, the conversation structure can be summarized as:
\begin{align*}
  \mathtt{Human:}     & \mathtt{<SYS>\ <TOKEN_{img}> <TOKEN_{act}>\ <STOP>} \\
  \mathtt{Agent:} & \mathtt{<TOKEN_{act}>\ <STOP>}
\end{align*}
where $\mathtt{<SYS>}$ is the system prompt and $\mathtt{<STOP>}$ is the stop token.

\noindent \textbf{Prompt for VDM.} 
The text encoder of VDM does not have the equivalent capacity for reasoning as that of the LLM. It struggles to comprehend the vehicle's right turn when the steer angle value is less than 0, while left turn when the value is greater than 0. To address this problem, we use GPT3.5~\cite{gpt3.5} to convert the low-level control signal into a motion description. As shown in Fig~\ref{fig:gpt}, the control signals from consecutive frames are used as the inputs. To generate clear motion prompts, we guide it to output common driving states such as steady speed, accelerating, decelerating, and turning. Conditioned on the motion prompts, the VDM is employed to predict the future frames.





\subsection{Model Training}
\label{sec:training}

As mentioned above, the overall architecture of ADriver-I includes the multi-modal large language model and video diffusion model. These two parts are trained separately and merged together during inference. Here, we will describe the training process of them.

\noindent \textbf{Training for MLLM:}
The MLLM is pretrained on our private dataset. It contains nearly 1.4M vision-action pairs on the highway scenario. 
For pretraining, we freeze the LLM model while the parameters of vision encoder and vision adapter layer are updated.
For supervised finetuning (SFT), we only freeze the vision encoder and train the rest parts. The SFT process is conducted on the nuScenes and our private datasets, respectively.
 

\noindent \textbf{Training for VDM:}
Following the training scheme in ~\cite{blattmann2023align}, the video diffusion model inherits the weights of stable diffusion~\cite{DBLP:conf/cvpr/RombachBLEO22} and is first pretrained on our 1.4M private dataset. Then the VDM is finetuned on the nuScenes dataset with about 23K video samples.




\begin{figure*}[t]
  \centering
   \includegraphics[width=1.0\linewidth]{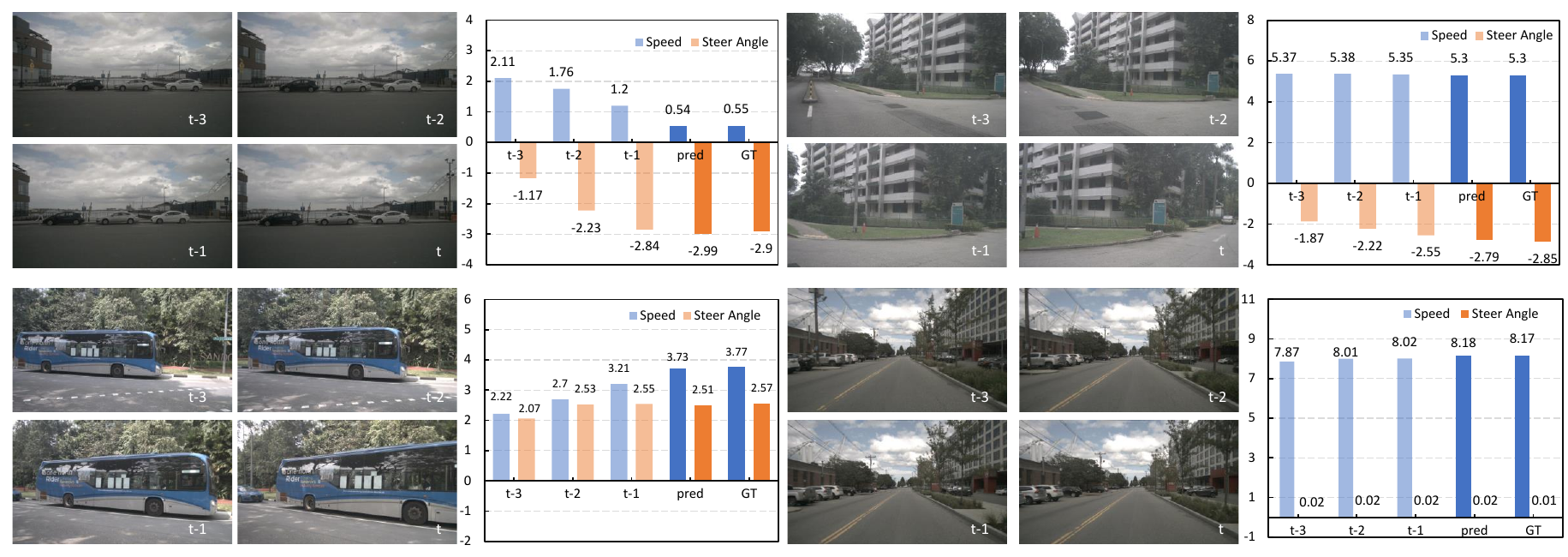}
   \vspace{-15pt}
   \caption{Qualitative visualization of control signal prediction. The left side shows input frames of timestamp $t-3$ to $t$. The right side's bar chart displays the actions for timestamp $t-3$ to $t-1$, with prediction (pred) and ground-truth (GT) at time $t$.}
   \label{fig:control signal}
\end{figure*}

\section{Experiments}
\label{sec:exp}

\subsection{Implementation Details}

\noindent \textbf{Multi-modal Large Language Model:}
MLLM is trained with 2 epochs for both pretraining and Supervised Fine-tuning with a batch size of 16. The input image size is $336 \times 336$. We take three historical vision-action pairs and current image as inputs. For the control signal, the number of decimal places is set to 3. To reduce the convergence difficulty of LLM, we multiply the digital number by 1000 to convert them into integer.
All the experiments are conducted on 8 A100 (80GB) GPUs. The AdamW optimizer is employed with the learning rate $2 \times 10^{-5}$. Considering the output of MLLM solely consists of text corpora, we follow LLaVA~\cite{llava} and use the cross-entropy loss for supervision.

\noindent \textbf{Video Diffusion Model:}
VDM is pretrained on the private dataset for 40k steps, utilizing a batch size of 128 across 32 A100 (80GB) GPUs. For fine-tuning, it is trained on the nuScenes dataset for an additional 40,000 steps with a reduced batch size of 32, using 16 A100 (80GB) GPUs. The spatial resolution is $256 \times 512$ and the video length is 8 both for pretraining and fine-tuneing. The learning rate is 4e-4 for pretraining and 3.2e-4 for fine-tuning, respectively. During inference, DDIM \cite{21DDIM} sampler is used and the sampling step is 50.

\subsection{Evaluation Metrics}
In order to quantitatively evaluate the control signal prediction and future scene generation, we adopt two evaluation metrics for these two tasks, respectively. To comprehensively evaluate the performance of control signal prediction, we employ the $L1$ metric to calculate the error between prediction and ground truth. We also compute the accuracy with different thresholds ($A_{\theta}$) for comparison. 
\begin{equation}
  F(\hat{x_i},x_i,\theta) = 
      \begin{cases} 
        1, \quad |\hat{x_i} - x_i| <= \theta  \\
        0, \quad |\hat{x_i} - x_i| > \theta 
       \end{cases} \\       
\label{eq:1}
\end{equation}
\begin{equation}
A_{\theta} = \frac{1}{N} \sum_{i =0}^{N} F(\hat{x_i},x_i,\theta),
\label{eq:1}
\end{equation}
where $N$ is the number of the validation samples and $\theta \in \{0.01, 0.03, 0.05, 0.07 \}$ is the threshold. $\hat{x_i}$ is the prediction and $x_i$ is the ground truth.

For the evaluation on quality of future generation, we employ frame-wise Frechet Inception Distance (FID)~\cite{heusel2017gans} and Frechet Video Distance (FVD)~\cite{unterthiner2018towards} as the metrics.

\begin{table}[t]
    \centering
    \begin{tabular}{c|cc}
        \toprule
         Number Embedding & Speed (m/s) & Steer angle (rad) \\
         \midrule
          Num2English & 2.094 & 0.536 \\
          Special Token & 0.094 & 0.106 \\
          Relative Diff & 0.081 & 0.096 \\
          \rowcolor[gray]{.9} Absolute Number & \textbf{0.072} & \textbf{0.091} \\
        \bottomrule
    \end{tabular}
    \caption{Encoding methods for control signal. ``Num2English" means translating the number into English expression. ``Special Token" divides the integer into several bins and converts it into classification problem. ``Relative Diff" represents the difference within the adjacent frames.}
    \label{tab:encodenum}
\end{table}

\subsection{Control Signal Prediction}

\noindent \textbf{Quantitative Results}
To show the effectiveness of our ADriver-I, we conduct experiments on nuScenes and our private datasets. As shown in Tab.~\ref{tab:quantitative}, we evaluate the performance of control signal prediction using the L1 error and accuracy with different thresholds. For cross-model comparison, we construct three competitive baselines based on the MLP, CNN and vision transformer (see Fig.~\ref{fig:multi-baseline} for details). The baseline (a) takes the historical action sequences of three frames as inputs and uses three fully-connected layers to predict the action of current frame. Based on the (a), baseline (b) further encodes the image features by a CNN (e.g., ResNet~\cite{he2016resnet}) and global average pooling (GAP). The image features are further concatenated with the action features and used to predict the current action. The baseline (c) simply replaces the CNN with a vision transformer (e.g., ViT-B~\cite{dosovitskiy2020image}) and shares the same design. The experimental results show that our ADriver-I outperforms these three baselines on nuScenes dataset. 
Fig.~\ref{fig:control signal} shows some qualitative visualizations of control signal prediction, given historical three frames.
However, we find that baseline (c) achieves similar performance on L1 error of steer angle, compared to ADriver-I. We infer some special cases with large steer angle variance contribute the most to the mean L1 error. The large advantage on accuracy with lower threshold $A_{0.01}$ compared to baselines support our opinion.

\begin{figure*}[t]
  \centering
   \includegraphics[width=1.0\linewidth]{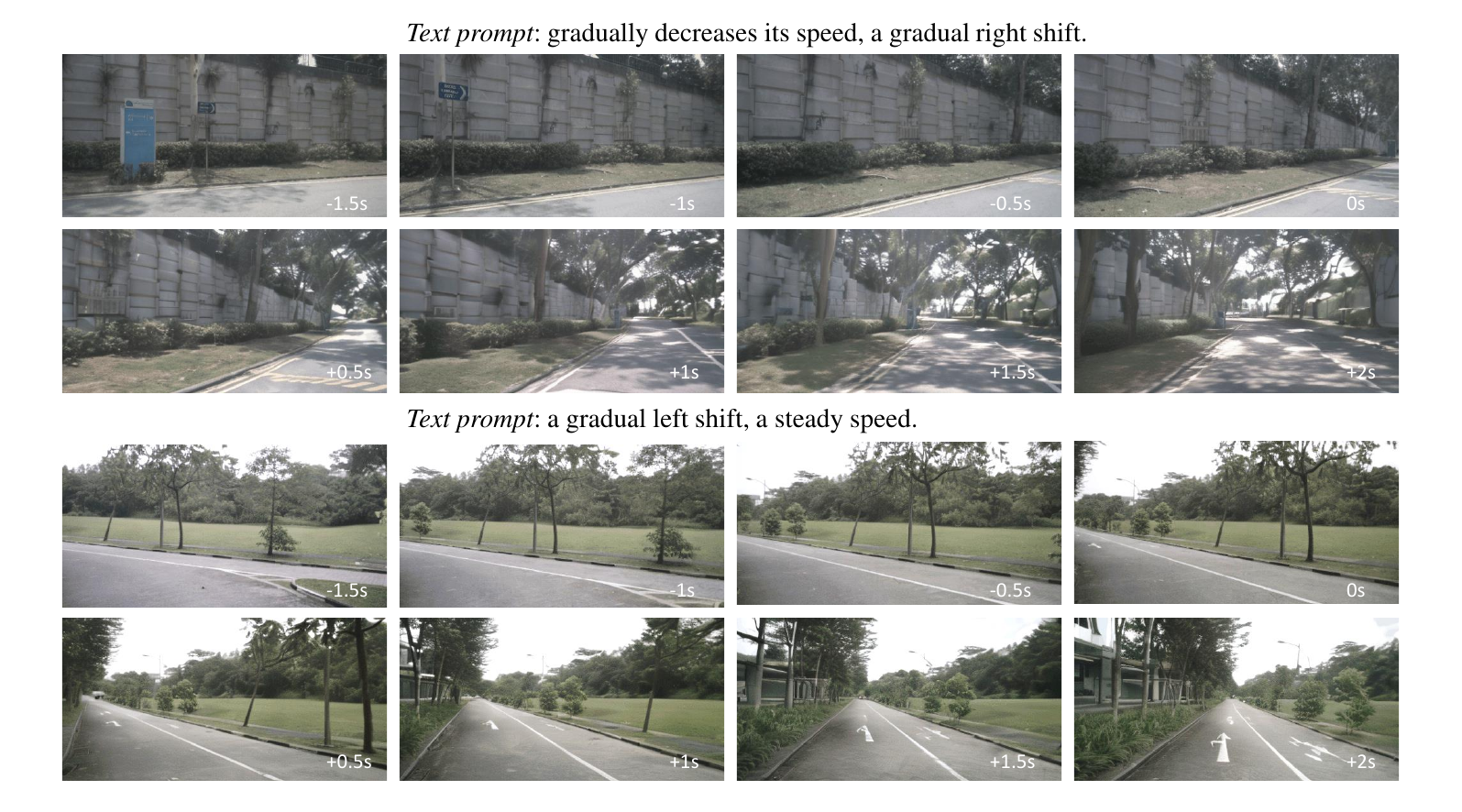}
   \vspace{-20pt}
   \caption{Qualitative visualization for future scene prediction. The first four images are the input historical four frames, while the latter four images show the predicted future scenes.}
   \label{fig:future prediction}
\end{figure*}

For cross-dataset comparison, our ADriver-I achieves impressive performance on our private dataset, compared to the nuScenes results. 
For example, it achieves L1 errors with 0.035 m/s and 0.015 rad on speed and steer angle prediction, respectively. There are two reasons to explain such phenomenon. The first one is that our private dataset mainly focus on the highway scene, where the speed and steer angle usually have relatively smaller variance. Also, its data scale for supervised finetuning is much larger than nuScenes (1.4M vs. 23K). Therefore, the performance of ADriver-I on private dataset is much better than nuScenes.

\begin{table}[t]
    \centering
    \vspace{-5pt}
    \begin{tabular}{c|cc}
        \toprule
         Decimal places & Speed (m/s) & Steer angle  (rad)\\
         \midrule
          0 & 0.212 & 0.099 \\
          1 & 0.094 & 0.093 \\
          2 & 0.073 & 0.091 \\
          \rowcolor[gray]{.9} 3 & \textbf{0.072} & \textbf{0.091} \\
        \bottomrule
    \end{tabular}
    \caption{Number of decimal places. Two decimal places produces almost the same performance compared to three decimal places.}
    \label{tab:numofdigital}
\end{table}

\noindent \textbf{Ablation Study}
We also conduct the ablation study to ablate some key designs for the control signal prediction. As shown in Tab.~\ref{tab:encodenum}, we first explore the encoding methods for control signal. It shows that directly predicting the absolute number of speed or steer angle produces better performance compared to the other courtparts, includes translating the number into English description, using relative difference. We also analyze the effect of the number of decimal places on the performance (see Tab.~\ref{tab:numofdigital}). The experiment results indicate that using two decimal places achieve similar performance compared to three decimal places, while outperforming fewer decimal places. It means integer or one decimal place tend to introduce some accuracy error. Finally, we further explore the effectiveness of our introduced multi-round conversation, refer to  Tab.~\ref{tab:multi-round}. The multi-round conversation outperforms the single-round one by a large margin on speed prediction. The multi-round conversation introduces more supervisions on the intermediate action predictions ($A^{'}_{t-2}$, $A^{'}_{t-1}$) during training. It greatly reduces the cumulative error on the action prediction $A^{'}_{t}$ of current frame during inference.

\begin{table}[t]
    \centering
    \vspace{-5pt}
    \begin{tabular}{c|cc}
        \toprule
         Conversation & Speed (m/s)& Steer angle  (rad)\\
         \midrule
          Temporal Fusion & 0.078 & 0.092 \\
          Single Round & 0.078 & 0.094\\
          \rowcolor[gray]{.9} Multi Round & \textbf{0.072} & \textbf{0.091}\\
        \bottomrule
    \end{tabular}
    \caption{Effectiveness of multi-round conversations. Temporal Fusion merges the multi-frame visual and action tokens. Single Round means only adding the supervision on current frame. Multi Round adds the supervision on all frames.}
    \label{tab:multi-round}
\end{table}

\begin{figure*}[t]
  \centering
   \includegraphics[width=1.0\linewidth]{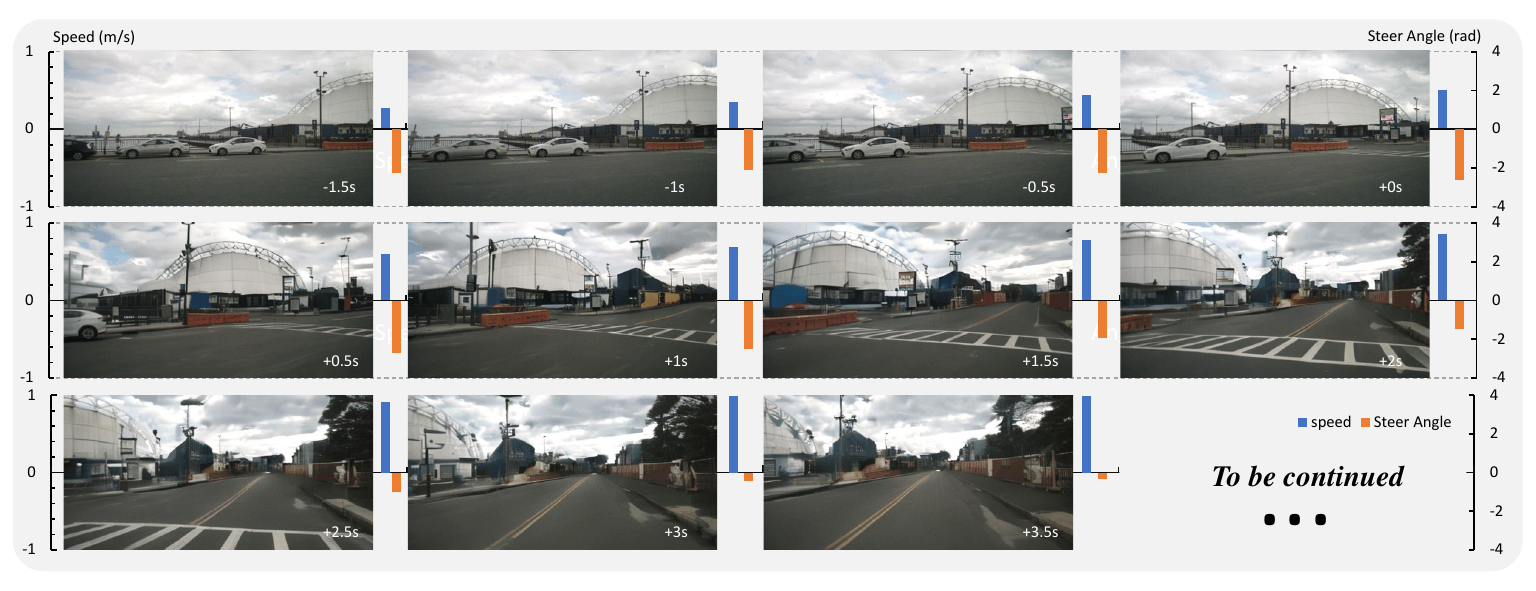}
   \vspace{-20pt}
   \caption{Qualitative visualization for long-term driving. Our \DriveLVLM{} achieves the autonomous driving in the world it creates.}
   \label{fig:infinite driving}
\end{figure*}

\subsection{Future Scene Generation}
\noindent \textbf{Qualitative Results}
Fig.~\ref{fig:future prediction} shows some qualitative results for future prediction generated by the video diffusion model. It provides two generation cases: turn left and turn right. The first and third rows represent the original video frames used as the condition and the second and fourth rows are the video frames predicted by the diffusion model. With the condition of historical frames and the control signal guidance predicted by MLLM, VDM can accordingly generate the future scenes without rely on any high-level knowledge such as the 3D bounding boxes and HD map.

\noindent \textbf{Quantitative Results}
To evaluate the quality of predicted future frames quantitatively, we calculate the FID and FVD metrics and report them in Tab.~\ref{tab:fidfvd} for reference. Without the 3D bounding boxes and HD maps as the input priors, our ADriver-I predicts the future four frames conditioned on historical four frames and achieves impressive performance with 5.5 FID and 97.0 FVD.

\subsection{Joint Control\&Generation}
As mentioned above,  ADriver-I has the possibility to drive in the world created by itself. To verify it, we only provide three historical interleaved vision-action pairs to ADriver-I and ADriver-I performs control signal prediction and future scene generation in a recurrent manner. As shown in Fig.~\ref{fig:infinite driving}, all the video frames are generated by the video diffusion model. The speed ({\color{blue1}blue bar}) and steer angle ({\color{orange1}orange bar}) predictions are produced by the MLLM. It shows that the predicted control signal can directly affect the future scene generation while the future scene generated pushes ADriver-I to take corresponding actions.

\begin{table}[t]
    \centering
    \begin{tabular}{c|c|cc}
        \toprule
         Method & Input$\rightarrow$Output & FID$\downarrow$ & FVD$\downarrow$ \\
         \midrule
          DriveGAN &1F$\rightarrow$12F & 73.4 & 502.3 \\
          DriveDreamer & 1F+12B+12M$\rightarrow$12F & 52.6 & 452.0 \\
          \rowcolor[gray]{.9} \DriveLVLM{} &4F $\rightarrow$ 4F & 5.5 & 97.0\\
        \bottomrule
    \end{tabular}
    \caption{Video generation performance comparison on nuScenes. ``F" denotes the frame. ``B" is the bouncing box and ``M" is the HD map. Note that all methods are conditioned on control signal.}
    \label{tab:fidfvd}
\end{table}


\section{Conclusion and Discussion}
\label{sec:conclusion}
In this paper, we construct a world model for autonomous driving called \DriveLVLM{}. It integrates the MLLM with VDM in a reasonable manner. It can directly output the low-level control signals based on current frame and historical vision-control pairs, without the need of mainstream sequential pipeline. It can also predict the near future frames based on historical information. Impressive performance is achieved on nuScenes and our private dataset. Given some initial vision-action pairs, \DriveLVLM{} achieves the infinite driving in the world generated by itself.

\noindent \textbf{Discussion:} 
The infinite driving achieved by \DriveLVLM{} provides some possibility for autonomous driving. 
There are still some drawbacks and limitations as follows:

(1) The generation module (VDM) is similar to the close-loop simulator, creating some unknown scenario for the MLLM to predict the corresponding control signals. However, VDM may generate some low-quality video frames especially when the control signals change quickly, disturbing the control signal prediction in the next timestamp.
(2) The performance is still far from satisfactory for deployment, we will update the \DriveLVLM{} version by training it with huge number of vision-control pairs in our increasing large-scale private dataset. 
(3) One drawback is that the MLLM and VDM are trained separately and fails to enjoy the benefits from end-to-end optimization. A unified comprehension\&generation framework is required to achieve this goal.
(4) Moreover, from the perspective of driving distance, there lacks of the routing information, we may introduce the navigation map to achieve the long-distance autonomous driving.

In summary, there is still a long road to go for the world model in autonomous driving. We hope to observe the scaling law from the generative perspective in the future.


{
    \small
    \bibliographystyle{ieeenat_fullname}
    \bibliography{main}

\begin{thebibliography}{67}
\providecommand{\natexlab}[1]{#1}
\providecommand{\url}[1]{\texttt{#1}}
\expandafter\ifx\csname urlstyle\endcsname\relax
  \providecommand{\doi}[1]{doi: #1}\else
  \providecommand{\doi}{doi: \begingroup \urlstyle{rm}\Url}\fi

\bibitem[Anonymous()]{Anonymous}
Anonymous.
\newblock Panacea: Panoramic and controllable video generation for autonomous driving.

\bibitem[Blattmann et~al.(2023)Blattmann, Rombach, Ling, Dockhorn, Kim, Fidler, and Kreis]{blattmann2023align}
Andreas Blattmann, Robin Rombach, Huan Ling, Tim Dockhorn, Seung~Wook Kim, Sanja Fidler, and Karsten Kreis.
\newblock Align your latents: High-resolution video synthesis with latent diffusion models.
\newblock In \emph{Proceedings of the IEEE/CVF Conference on Computer Vision and Pattern Recognition}, pages 22563--22575, 2023.

\bibitem[Brohan et~al.(2023{\natexlab{a}})Brohan, Brown, Carbajal, Chebotar, Chen, Choromanski, Ding, Driess, Dubey, Finn, Florence, Fu, Arenas, Gopalakrishnan, Han, Hausman, Herzog, Hsu, Ichter, Irpan, Joshi, Julian, Kalashnikov, Kuang, Leal, Lee, Lee, Levine, Lu, Michalewski, Mordatch, Pertsch, Rao, Reymann, Ryoo, Salazar, Sanketi, Sermanet, Singh, Singh, Soricut, Tran, Vanhoucke, Vuong, Wahid, Welker, Wohlhart, Wu, Xia, Xiao, Xu, Xu, Yu, and Zitkovich]{RT-2}
Anthony Brohan, Noah Brown, Justice Carbajal, Yevgen Chebotar, Xi Chen, Krzysztof Choromanski, Tianli Ding, Danny Driess, Avinava Dubey, Chelsea Finn, Pete Florence, Chuyuan Fu, Montse~Gonzalez Arenas, Keerthana Gopalakrishnan, Kehang Han, Karol Hausman, Alexander Herzog, Jasmine Hsu, Brian Ichter, Alex Irpan, Nikhil~J. Joshi, Ryan Julian, Dmitry Kalashnikov, Yuheng Kuang, Isabel Leal, Lisa Lee, Tsang{-}Wei~Edward Lee, Sergey Levine, Yao Lu, Henryk Michalewski, Igor Mordatch, Karl Pertsch, Kanishka Rao, Krista Reymann, Michael~S. Ryoo, Grecia Salazar, Pannag Sanketi, Pierre Sermanet, Jaspiar Singh, Anikait Singh, Radu Soricut, Huong~T. Tran, Vincent Vanhoucke, Quan Vuong, Ayzaan Wahid, Stefan Welker, Paul Wohlhart, Jialin Wu, Fei Xia, Ted Xiao, Peng Xu, Sichun Xu, Tianhe Yu, and Brianna Zitkovich.
\newblock {RT-2:} vision-language-action models transfer web knowledge to robotic control.
\newblock \emph{CoRR}, abs/2307.15818, 2023{\natexlab{a}}.

\bibitem[Brohan et~al.(2023{\natexlab{b}})Brohan, Brown, Carbajal, Chebotar, Dabis, Finn, Gopalakrishnan, Hausman, Herzog, Hsu, Ibarz, Ichter, Irpan, Jackson, Jesmonth, Joshi, Julian, Kalashnikov, Kuang, Leal, Lee, Levine, Lu, Malla, Manjunath, Mordatch, Nachum, Parada, Peralta, Perez, Pertsch, Quiambao, Rao, Ryoo, Salazar, Sanketi, Sayed, Singh, Sontakke, Stone, Tan, Tran, Vanhoucke, Vega, Vuong, Xia, Xiao, Xu, Xu, Yu, and Zitkovich]{RT-1}
Anthony Brohan, Noah Brown, Justice Carbajal, Yevgen Chebotar, Joseph Dabis, Chelsea Finn, Keerthana Gopalakrishnan, Karol Hausman, Alexander Herzog, Jasmine Hsu, Julian Ibarz, Brian Ichter, Alex Irpan, Tomas Jackson, Sally Jesmonth, Nikhil~J. Joshi, Ryan Julian, Dmitry Kalashnikov, Yuheng Kuang, Isabel Leal, Kuang{-}Huei Lee, Sergey Levine, Yao Lu, Utsav Malla, Deeksha Manjunath, Igor Mordatch, Ofir Nachum, Carolina Parada, Jodilyn Peralta, Emily Perez, Karl Pertsch, Jornell Quiambao, Kanishka Rao, Michael~S. Ryoo, Grecia Salazar, Pannag~R. Sanketi, Kevin Sayed, Jaspiar Singh, Sumedh Sontakke, Austin Stone, Clayton Tan, Huong~T. Tran, Vincent Vanhoucke, Steve Vega, Quan Vuong, Fei Xia, Ted Xiao, Peng Xu, Sichun Xu, Tianhe Yu, and Brianna Zitkovich.
\newblock {RT-1:} robotics transformer for real-world control at scale.
\newblock In \emph{Robotics: Science and Systems XIX, Daegu, Republic of Korea, July 10-14, 2023}, 2023{\natexlab{b}}.

\bibitem[Chiang et~al.(2023)Chiang, Li, Lin, Sheng, Wu, Zhang, Zheng, Zhuang, Zhuang, Gonzalez, et~al.]{vicuna1.5}
Wei-Lin Chiang, Zhuohan Li, Zi Lin, Ying Sheng, Zhanghao Wu, Hao Zhang, Lianmin Zheng, Siyuan Zhuang, Yonghao Zhuang, Joseph~E Gonzalez, et~al.
\newblock Vicuna: An open-source chatbot impressing gpt-4 with 90\%* chatgpt quality.
\newblock \emph{See https://vicuna. lmsys. org (accessed 14 April 2023)}, 2023.

\bibitem[Chitta et~al.(2021)Chitta, Prakash, and Geiger]{chitta2021neat}
Kashyap Chitta, Aditya Prakash, and Andreas Geiger.
\newblock Neat: Neural attention fields for end-to-end autonomous driving.
\newblock In \emph{Proceedings of the IEEE/CVF International Conference on Computer Vision}, pages 15793--15803, 2021.

\bibitem[Chowdhery et~al.(2023)Chowdhery, Narang, Devlin, Bosma, Mishra, Roberts, Barham, Chung, Sutton, Gehrmann, Schuh, Shi, Tsvyashchenko, Maynez, Rao, Barnes, Tay, Shazeer, Prabhakaran, Reif, Du, Hutchinson, Pope, Bradbury, Austin, Isard, Gur{-}Ari, Yin, Duke, Levskaya, Ghemawat, Dev, Michalewski, Garcia, Misra, Robinson, Fedus, Zhou, Ippolito, Luan, Lim, Zoph, Spiridonov, Sepassi, Dohan, Agrawal, Omernick, Dai, Pillai, Pellat, Lewkowycz, Moreira, Child, Polozov, Lee, Zhou, Wang, Saeta, Diaz, Firat, Catasta, Wei, Meier{-}Hellstern, Eck, Dean, Petrov, and Fiedel]{chowdhery2022palm}
Aakanksha Chowdhery, Sharan Narang, Jacob Devlin, Maarten Bosma, Gaurav Mishra, Adam Roberts, Paul Barham, Hyung~Won Chung, Charles Sutton, Sebastian Gehrmann, Parker Schuh, Kensen Shi, Sasha Tsvyashchenko, Joshua Maynez, Abhishek Rao, Parker Barnes, Yi Tay, Noam Shazeer, Vinodkumar Prabhakaran, Emily Reif, Nan Du, Ben Hutchinson, Reiner Pope, James Bradbury, Jacob Austin, Michael Isard, Guy Gur{-}Ari, Pengcheng Yin, Toju Duke, Anselm Levskaya, Sanjay Ghemawat, Sunipa Dev, Henryk Michalewski, Xavier Garcia, Vedant Misra, Kevin Robinson, Liam Fedus, Denny Zhou, Daphne Ippolito, David Luan, Hyeontaek Lim, Barret Zoph, Alexander Spiridonov, Ryan Sepassi, David Dohan, Shivani Agrawal, Mark Omernick, Andrew~M. Dai, Thanumalayan~Sankaranarayana Pillai, Marie Pellat, Aitor Lewkowycz, Erica Moreira, Rewon Child, Oleksandr Polozov, Katherine Lee, Zongwei Zhou, Xuezhi Wang, Brennan Saeta, Mark Diaz, Orhan Firat, Michele Catasta, Jason Wei, Kathy Meier{-}Hellstern, Douglas Eck, Jeff Dean, Slav Petrov, and Noah Fiedel.
\newblock Palm: Scaling language modeling with pathways.
\newblock \emph{J. Mach. Learn. Res.}, 2023.

\bibitem[Dehghani et~al.(2023)Dehghani, Djolonga, Mustafa, Padlewski, Heek, Gilmer, Steiner, Caron, Geirhos, Alabdulmohsin, Jenatton, Beyer, Tschannen, Arnab, Wang, Ruiz, Minderer, Puigcerver, Evci, Kumar, van Steenkiste, Elsayed, Mahendran, Yu, Oliver, Huot, Bastings, Collier, Gritsenko, Birodkar, Vasconcelos, Tay, Mensink, Kolesnikov, Pavetic, Tran, Kipf, Lucic, Zhai, Keysers, Harmsen, and Houlsby]{2023vit22b}
Mostafa Dehghani, Josip Djolonga, Basil Mustafa, Piotr Padlewski, Jonathan Heek, Justin Gilmer, Andreas~Peter Steiner, Mathilde Caron, Robert Geirhos, Ibrahim Alabdulmohsin, Rodolphe Jenatton, Lucas Beyer, Michael Tschannen, Anurag Arnab, Xiao Wang, Carlos~Riquelme Ruiz, Matthias Minderer, Joan Puigcerver, Utku Evci, Manoj Kumar, Sjoerd van Steenkiste, Gamaleldin~Fathy Elsayed, Aravindh Mahendran, Fisher Yu, Avital Oliver, Fantine Huot, Jasmijn Bastings, Mark Collier, Alexey~A. Gritsenko, Vighnesh Birodkar, Cristina~Nader Vasconcelos, Yi Tay, Thomas Mensink, Alexander Kolesnikov, Filip Pavetic, Dustin Tran, Thomas Kipf, Mario Lucic, Xiaohua Zhai, Daniel Keysers, Jeremiah~J. Harmsen, and Neil Houlsby.
\newblock Scaling vision transformers to 22 billion parameters.
\newblock pages 7480--7512. {PMLR}, 2023.

\bibitem[Dosovitskiy et~al.(2020)Dosovitskiy, Beyer, Kolesnikov, Weissenborn, Zhai, Unterthiner, Dehghani, Minderer, Heigold, Gelly, et~al.]{dosovitskiy2020image}
Alexey Dosovitskiy, Lucas Beyer, Alexander Kolesnikov, Dirk Weissenborn, Xiaohua Zhai, Thomas Unterthiner, Mostafa Dehghani, Matthias Minderer, Georg Heigold, Sylvain Gelly, et~al.
\newblock An image is worth 16x16 words: Transformers for image recognition at scale.
\newblock \emph{arXiv preprint arXiv:2010.11929}, 2020.

\bibitem[Driess et~al.(2023)Driess, Xia, Sajjadi, Lynch, Chowdhery, Ichter, Wahid, Tompson, Vuong, Yu, Huang, Chebotar, Sermanet, Duckworth, Levine, Vanhoucke, Hausman, Toussaint, Greff, Zeng, Mordatch, and Florence]{PaLM-E}
Danny Driess, Fei Xia, Mehdi S.~M. Sajjadi, Corey Lynch, Aakanksha Chowdhery, Brian Ichter, Ayzaan Wahid, Jonathan Tompson, Quan Vuong, Tianhe Yu, Wenlong Huang, Yevgen Chebotar, Pierre Sermanet, Daniel Duckworth, Sergey Levine, Vincent Vanhoucke, Karol Hausman, Marc Toussaint, Klaus Greff, Andy Zeng, Igor Mordatch, and Pete Florence.
\newblock Palm-e: An embodied multimodal language model.
\newblock In \emph{International Conference on Machine Learning, {ICML} 2023, 23-29 July 2023, Honolulu, Hawaii, {USA}}. {PMLR}, 2023.

\bibitem[Du et~al.(2022)Du, Qian, Liu, Ding, Qiu, Yang, and Tang]{du2022glm}
Zhengxiao Du, Yujie Qian, Xiao Liu, Ming Ding, Jiezhong Qiu, Zhilin Yang, and Jie Tang.
\newblock Glm: General language model pretraining with autoregressive blank infilling.
\newblock In \emph{Proceedings of the 60th Annual Meeting of the Association for Computational Linguistics (Volume 1: Long Papers)}, pages 320--335, 2022.

\bibitem[Goodfellow et~al.(2014)Goodfellow, Pouget-Abadie, Mirza, Xu, Warde-Farley, Ozair, Courville, and Bengio]{goodfellow2014GAN}
Ian Goodfellow, Jean Pouget-Abadie, Mehdi Mirza, Bing Xu, David Warde-Farley, Sherjil Ozair, Aaron Courville, and Yoshua Bengio.
\newblock Generative adversarial nets.
\newblock \emph{Advances in neural information processing systems}, 27, 2014.

\bibitem[Ha and Schmidhuber(2018)]{HaS18Recurrentworld}
David Ha and J{\"{u}}rgen Schmidhuber.
\newblock Recurrent world models facilitate policy evolution.
\newblock In \emph{Advances in Neural Information Processing Systems 31: Annual Conference on Neural Information Processing Systems 2018, NeurIPS 2018, December 3-8, 2018, Montr{\'{e}}al, Canada}, pages 2455--2467, 2018.

\bibitem[Hafner et~al.(2020)Hafner, Lillicrap, Ba, and Norouzi]{20DreamtoControl}
Danijar Hafner, Timothy~P. Lillicrap, Jimmy Ba, and Mohammad Norouzi.
\newblock Dream to control: Learning behaviors by latent imagination.
\newblock In \emph{8th International Conference on Learning Representations, {ICLR} 2020, Addis Ababa, Ethiopia, April 26-30, 2020}. OpenReview.net, 2020.

\bibitem[He et~al.(2016)He, Zhang, Ren, and Sun]{he2016resnet}
Kaiming He, Xiangyu Zhang, Shaoqing Ren, and Jian Sun.
\newblock Deep residual learning for image recognition.
\newblock In \emph{Proceedings of the IEEE conference on computer vision and pattern recognition}, pages 770--778, 2016.

\bibitem[He et~al.(2017)He, Gkioxari, Doll{\'{a}}r, and Girshick]{2017maskrcnn}
Kaiming He, Georgia Gkioxari, Piotr Doll{\'{a}}r, and Ross~B. Girshick.
\newblock Mask {R-CNN}.
\newblock In \emph{ICCV}. {IEEE} Computer Society, 2017.

\bibitem[Heusel et~al.(2017)Heusel, Ramsauer, Unterthiner, Nessler, and Hochreiter]{heusel2017gans}
Martin Heusel, Hubert Ramsauer, Thomas Unterthiner, Bernhard Nessler, and Sepp Hochreiter.
\newblock Gans trained by a two time-scale update rule converge to a local nash equilibrium.
\newblock \emph{Advances in neural information processing systems}, 30, 2017.

\bibitem[Ho et~al.(2020)Ho, Jain, and Abbeel]{20DDPM}
Jonathan Ho, Ajay Jain, and Pieter Abbeel.
\newblock Denoising diffusion probabilistic models.
\newblock In \emph{Advances in Neural Information Processing Systems 33: Annual Conference on Neural Information Processing Systems 2020, NeurIPS 2020, December 6-12, 2020, virtual}, 2020.

\bibitem[Hu et~al.(2023{\natexlab{a}})Hu, Russell, Yeo, Murez, Fedoseev, Kendall, Shotton, and Corrado]{hu2023GAIA}
Anthony Hu, Lloyd Russell, Hudson Yeo, Zak Murez, George Fedoseev, Alex Kendall, Jamie Shotton, and Gianluca Corrado.
\newblock Gaia-1: A generative world model for autonomous driving.
\newblock \emph{arXiv preprint arXiv:2309.17080}, 2023{\natexlab{a}}.

\bibitem[Hu et~al.(2022)Hu, Chen, Wu, Li, Yan, and Tao]{hu2022st}
Shengchao Hu, Li Chen, Penghao Wu, Hongyang Li, Junchi Yan, and Dacheng Tao.
\newblock St-p3: End-to-end vision-based autonomous driving via spatial-temporal feature learning.
\newblock In \emph{European Conference on Computer Vision}, pages 533--549. Springer, 2022.

\bibitem[Hu et~al.(2023{\natexlab{b}})Hu, Yang, Chen, Li, Sima, Zhu, Chai, Du, Lin, Wang, et~al.]{hu2023uniad}
Yihan Hu, Jiazhi Yang, Li Chen, Keyu Li, Chonghao Sima, Xizhou Zhu, Siqi Chai, Senyao Du, Tianwei Lin, Wenhai Wang, et~al.
\newblock Planning-oriented autonomous driving.
\newblock In \emph{Proceedings of the IEEE/CVF Conference on Computer Vision and Pattern Recognition}, pages 17853--17862, 2023{\natexlab{b}}.

\bibitem[Huang et~al.(2023)Huang, Wang, Zhang, Li, Wu, and Fei-Fei]{huang2023voxposer}
Wenlong Huang, Chen Wang, Ruohan Zhang, Yunzhu Li, Jiajun Wu, and Li Fei-Fei.
\newblock Voxposer: Composable 3d value maps for robotic manipulation with language models.
\newblock \emph{arXiv preprint arXiv:2307.05973}, 2023.

\bibitem[Jiang et~al.(2023)Jiang, Chen, Xu, Liao, Chen, Zhou, Zhang, Liu, Huang, and Wang]{jiang2023vad}
Bo Jiang, Shaoyu Chen, Qing Xu, Bencheng Liao, Jiajie Chen, Helong Zhou, Qian Zhang, Wenyu Liu, Chang Huang, and Xinggang Wang.
\newblock Vad: Vectorized scene representation for efficient autonomous driving.
\newblock \emph{arXiv preprint arXiv:2303.12077}, 2023.

\bibitem[Jiang et~al.(2022)Jiang, Gupta, Zhang, Wang, Dou, Chen, Fei-Fei, Anandkumar, Zhu, and Fan]{jiang2022vima}
Yunfan Jiang, Agrim Gupta, Zichen Zhang, Guanzhi Wang, Yongqiang Dou, Yanjun Chen, Li Fei-Fei, Anima Anandkumar, Yuke Zhu, and Linxi Fan.
\newblock Vima: General robot manipulation with multimodal prompts.
\newblock \emph{arXiv}, 2022.

\bibitem[Kim et~al.(2020)Kim, Zhou, Philion, Torralba, and Fidler]{20GameGAN}
Seung~Wook Kim, Yuhao Zhou, Jonah Philion, Antonio Torralba, and Sanja Fidler.
\newblock Learning to simulate dynamic environments with gamegan.
\newblock In \emph{2020 {IEEE/CVF} Conference on Computer Vision and Pattern Recognition, {CVPR} 2020, Seattle, WA, USA, June 13-19, 2020}, 2020.

\bibitem[Kim et~al.(2021)Kim, Philion, Torralba, and Fidler]{21DriveGAN}
Seung~Wook Kim, Jonah Philion, Antonio Torralba, and Sanja Fidler.
\newblock Drivegan: Towards a controllable high-quality neural simulation.
\newblock In \emph{{IEEE} Conference on Computer Vision and Pattern Recognition, {CVPR} 2021, virtual, June 19-25, 2021}, 2021.

\bibitem[Kingma and Welling(2014)]{2014VAE}
Diederik~P. Kingma and Max Welling.
\newblock Auto-encoding variational bayes.
\newblock In \emph{2nd International Conference on Learning Representations, {ICLR} 2014, Banff, AB, Canada, April 14-16, 2014, Conference Track Proceedings}, 2014.

\bibitem[Le et~al.(2023)Le, Vyas, Shi, Karrer, Sari, Moritz, Williamson, Manohar, Adi, Mahadeokar, et~al.]{le2023voicebox}
Matthew Le, Apoorv Vyas, Bowen Shi, Brian Karrer, Leda Sari, Rashel Moritz, Mary Williamson, Vimal Manohar, Yossi Adi, Jay Mahadeokar, et~al.
\newblock Voicebox: Text-guided multilingual universal speech generation at scale.
\newblock \emph{arXiv preprint arXiv:2306.15687}, 2023.

\bibitem[Li et~al.(2023{\natexlab{a}})Li, Li, Savarese, and Hoi]{23blip2}
Junnan Li, Dongxu Li, Silvio Savarese, and Steven C.~H. Hoi.
\newblock {BLIP-2:} bootstrapping language-image pre-training with frozen image encoders and large language models.
\newblock In \emph{International Conference on Machine Learning, {ICML} 2023, 23-29 July 2023, Honolulu, Hawaii, {USA}}, 2023{\natexlab{a}}.

\bibitem[Li et~al.(2023{\natexlab{b}})Li, He, Wang, Li, Wang, Luo, Wang, Wang, and Qiao]{li2023videochat}
KunChang Li, Yinan He, Yi Wang, Yizhuo Li, Wenhai Wang, Ping Luo, Yali Wang, Limin Wang, and Yu Qiao.
\newblock Videochat: Chat-centric video understanding.
\newblock \emph{arXiv preprint arXiv:2305.06355}, 2023{\natexlab{b}}.

\bibitem[Liu et~al.(2023{\natexlab{a}})Liu, Li, Li, and Lee]{llava1.5}
Haotian Liu, Chunyuan Li, Yuheng Li, and Yong~Jae Lee.
\newblock Improved baselines with visual instruction tuning.
\newblock \emph{arXiv preprint arXiv:2310.03744}, 2023{\natexlab{a}}.

\bibitem[Liu et~al.(2023{\natexlab{b}})Liu, Li, Wu, and Lee]{llava}
Haotian Liu, Chunyuan Li, Qingyang Wu, and Yong~Jae Lee.
\newblock Visual instruction tuning.
\newblock \emph{arXiv preprint arXiv:2304.08485}, 2023{\natexlab{b}}.

\bibitem[Liu et~al.(2022{\natexlab{a}})Liu, Wang, Zhang, and Sun]{liu2022petr}
Yingfei Liu, Tiancai Wang, Xiangyu Zhang, and Jian Sun.
\newblock Petr: Position embedding transformation for multi-view 3d object detection.
\newblock \emph{arXiv preprint arXiv:2203.05625}, 2022{\natexlab{a}}.

\bibitem[Liu et~al.(2022{\natexlab{b}})Liu, Yan, Jia, Li, Gao, Wang, Zhang, and Sun]{liu2022petrv2}
Yingfei Liu, Junjie Yan, Fan Jia, Shuailin Li, Qi Gao, Tiancai Wang, Xiangyu Zhang, and Jian Sun.
\newblock Petrv2: A unified framework for 3d perception from multi-camera images.
\newblock \emph{arXiv preprint arXiv:2206.01256}, 2022{\natexlab{b}}.

\bibitem[Luo et~al.(2023)Luo, Zhao, Yang, Dong, Qiu, Lu, Wang, and Wei]{luo2023valley}
Ruipu Luo, Ziwang Zhao, Min Yang, Junwei Dong, Minghui Qiu, Pengcheng Lu, Tao Wang, and Zhongyu Wei.
\newblock Valley: Video assistant with large language model enhanced ability.
\newblock \emph{arXiv preprint arXiv:2306.07207}, 2023.

\bibitem[OpenAI(2023{\natexlab{a}})]{gpt3.5}
OpenAI.
\newblock \url{https://chat.openai.com}, 2023{\natexlab{a}}.

\bibitem[OpenAI(2023{\natexlab{b}})]{gpt4}
OpenAI.
\newblock Gpt-4 technical report, 2023{\natexlab{b}}.

\bibitem[OpenAI(2023{\natexlab{c}})]{gpt4v}
OpenAI.
\newblock Gpt-4v(ision) system card.
\newblock 2023{\natexlab{c}}.

\bibitem[Peng et~al.(2023)Peng, Li, He, Galley, and Gao]{peng2023vicuna}
Baolin Peng, Chunyuan Li, Pengcheng He, Michel Galley, and Jianfeng Gao.
\newblock Instruction tuning with gpt-4.
\newblock \emph{arXiv preprint arXiv:2304.03277}, 2023.

\bibitem[Prakash et~al.(2021)Prakash, Chitta, and Geiger]{prakash2021multi}
Aditya Prakash, Kashyap Chitta, and Andreas Geiger.
\newblock Multi-modal fusion transformer for end-to-end autonomous driving.
\newblock In \emph{Proceedings of the IEEE/CVF Conference on Computer Vision and Pattern Recognition}, pages 7077--7087, 2021.

\bibitem[Radford et~al.(2018)Radford, Narasimhan, Salimans, Sutskever, et~al.]{2018GPT1}
Alec Radford, Karthik Narasimhan, Tim Salimans, Ilya Sutskever, et~al.
\newblock Improving language understanding by generative pre-training.
\newblock 2018.

\bibitem[Radford et~al.(2019{\natexlab{a}})Radford, Wu, Child, Luan, Amodei, Sutskever, et~al.]{2019GPT2}
Alec Radford, Jeffrey Wu, Rewon Child, David Luan, Dario Amodei, Ilya Sutskever, et~al.
\newblock Language models are unsupervised multitask learners.
\newblock \emph{OpenAI blog}, 1\penalty0 (8):\penalty0 9, 2019{\natexlab{a}}.

\bibitem[Radford et~al.(2019{\natexlab{b}})Radford, Wu, Child, Luan, Amodei, Sutskever, et~al.]{2019GPT3}
Alec Radford, Jeffrey Wu, Rewon Child, David Luan, Dario Amodei, Ilya Sutskever, et~al.
\newblock Language models are unsupervised multitask learners.
\newblock \emph{OpenAI blog}, 1\penalty0 (8):\penalty0 9, 2019{\natexlab{b}}.

\bibitem[Radford et~al.(2021)Radford, Kim, Hallacy, Ramesh, Goh, Agarwal, Sastry, Askell, Mishkin, Clark, Krueger, and Sutskever]{CLIP}
Alec Radford, Jong~Wook Kim, Chris Hallacy, Aditya Ramesh, Gabriel Goh, Sandhini Agarwal, Girish Sastry, Amanda Askell, Pamela Mishkin, Jack Clark, Gretchen Krueger, and Ilya Sutskever.
\newblock Learning transferable visual models from natural language supervision.
\newblock In \emph{Proceedings of the 38th International Conference on Machine Learning, {ICML} 2021, 18-24 July 2021, Virtual Event}. {PMLR}, 2021.

\bibitem[Rombach et~al.(2022{\natexlab{a}})Rombach, Blattmann, Lorenz, Esser, and Ommer]{DBLP:conf/cvpr/RombachBLEO22}
Robin Rombach, Andreas Blattmann, Dominik Lorenz, Patrick Esser, and Bj{\"{o}}rn Ommer.
\newblock High-resolution image synthesis with latent diffusion models.
\newblock In \emph{{IEEE/CVF} Conference on Computer Vision and Pattern Recognition, {CVPR} 2022, New Orleans, LA, USA, June 18-24, 2022}, pages 10674--10685. {IEEE}, 2022{\natexlab{a}}.

\bibitem[Rombach et~al.(2022{\natexlab{b}})Rombach, Blattmann, Lorenz, Esser, and Ommer]{Rombach_2022_CVPR}
Robin Rombach, Andreas Blattmann, Dominik Lorenz, Patrick Esser, and Bj\"orn Ommer.
\newblock High-resolution image synthesis with latent diffusion models.
\newblock In \emph{Proceedings of the IEEE/CVF Conference on Computer Vision and Pattern Recognition (CVPR)}, pages 10684--10695, 2022{\natexlab{b}}.

\bibitem[Rombach et~al.(2022{\natexlab{c}})Rombach, Blattmann, Lorenz, Esser, and Ommer]{rombach2022high}
Robin Rombach, Andreas Blattmann, Dominik Lorenz, Patrick Esser, and Bj{\"o}rn Ommer.
\newblock High-resolution image synthesis with latent diffusion models.
\newblock In \emph{Proceedings of the IEEE/CVF conference on computer vision and pattern recognition}, pages 10684--10695, 2022{\natexlab{c}}.

\bibitem[Rubenstein et~al.(2023)Rubenstein, Asawaroengchai, Nguyen, Bapna, Borsos, Quitry, Chen, Badawy, Han, Kharitonov, et~al.]{rubenstein2023audiopalm}
Paul~K Rubenstein, Chulayuth Asawaroengchai, Duc~Dung Nguyen, Ankur Bapna, Zal{\'a}n Borsos, F{\'e}lix de~Chaumont Quitry, Peter Chen, Dalia~El Badawy, Wei Han, Eugene Kharitonov, et~al.
\newblock Audiopalm: A large language model that can speak and listen.
\newblock \emph{arXiv preprint arXiv:2306.12925}, 2023.

\bibitem[Seo et~al.(2023)Seo, Hafner, Liu, Liu, James, Lee, and Abbeel]{seo2023masked}
Younggyo Seo, Danijar Hafner, Hao Liu, Fangchen Liu, Stephen James, Kimin Lee, and Pieter Abbeel.
\newblock Masked world models for visual control.
\newblock In \emph{Conference on Robot Learning}, pages 1332--1344. PMLR, 2023.

\bibitem[Song et~al.(2021)Song, Meng, and Ermon]{21DDIM}
Jiaming Song, Chenlin Meng, and Stefano Ermon.
\newblock Denoising diffusion implicit models.
\newblock In \emph{9th International Conference on Learning Representations, {ICLR} 2021, Virtual Event, Austria, May 3-7, 2021}. OpenReview.net, 2021.

\bibitem[Swerdlow et~al.(2023)Swerdlow, Xu, and Zhou]{23bevgen}
Alexander Swerdlow, Runsheng Xu, and Bolei Zhou.
\newblock Street-view image generation from a bird's-eye view layout.
\newblock \emph{arXiv preprint arXiv:2301.04634}, 2023.

\bibitem[Touvron et~al.(2023{\natexlab{a}})Touvron, Lavril, Izacard, Martinet, Lachaux, Lacroix, Rozi{\`e}re, Goyal, Hambro, Azhar, et~al.]{touvron2023llama}
Hugo Touvron, Thibaut Lavril, Gautier Izacard, Xavier Martinet, Marie-Anne Lachaux, Timoth{\'e}e Lacroix, Baptiste Rozi{\`e}re, Naman Goyal, Eric Hambro, Faisal Azhar, et~al.
\newblock Llama: Open and efficient foundation language models.
\newblock \emph{arXiv preprint arXiv:2302.13971}, 2023{\natexlab{a}}.

\bibitem[Touvron et~al.(2023{\natexlab{b}})Touvron, Martin, Stone, Albert, Almahairi, Babaei, Bashlykov, Batra, Bhargava, Bhosale, et~al.]{touvron2023llama2}
Hugo Touvron, Louis Martin, Kevin Stone, Peter Albert, Amjad Almahairi, Yasmine Babaei, Nikolay Bashlykov, Soumya Batra, Prajjwal Bhargava, Shruti Bhosale, et~al.
\newblock Llama 2: Open foundation and fine-tuned chat models.
\newblock \emph{arXiv preprint arXiv:2307.09288}, 2023{\natexlab{b}}.

\bibitem[Unterthiner et~al.(2018)Unterthiner, Van~Steenkiste, Kurach, Marinier, Michalski, and Gelly]{unterthiner2018towards}
Thomas Unterthiner, Sjoerd Van~Steenkiste, Karol Kurach, Raphael Marinier, Marcin Michalski, and Sylvain Gelly.
\newblock Towards accurate generative models of video: A new metric \& challenges.
\newblock \emph{arXiv preprint arXiv:1812.01717}, 2018.

\bibitem[van~den Oord et~al.(2017)van~den Oord, Vinyals, and Kavukcuoglu]{van2017VQVAE}
A{\"{a}}ron van~den Oord, Oriol Vinyals, and Koray Kavukcuoglu.
\newblock Neural discrete representation learning.
\newblock In \emph{Advances in Neural Information Processing Systems 30: Annual Conference on Neural Information Processing Systems 2017, December 4-9, 2017, Long Beach, CA, {USA}}, pages 6306--6315, 2017.

\bibitem[Wang et~al.(2023{\natexlab{a}})Wang, Chen, Luo, Dai, Yuan, Wu, and Jiang]{wang2023chatvideo}
Junke Wang, Dongdong Chen, Chong Luo, Xiyang Dai, Lu Yuan, Zuxuan Wu, and Yu-Gang Jiang.
\newblock Chatvideo: A tracklet-centric multimodal and versatile video understanding system.
\newblock \emph{arXiv preprint arXiv:2304.14407}, 2023{\natexlab{a}}.

\bibitem[Wang et~al.(2023{\natexlab{b}})Wang, Liu, Wang, Li, and Zhang]{wang2023streampetr}
Shihao Wang, Yingfei Liu, Tiancai Wang, Ying Li, and Xiangyu Zhang.
\newblock Exploring object-centric temporal modeling for efficient multi-view 3d object detection.
\newblock \emph{arXiv preprint arXiv:2303.11926}, 2023{\natexlab{b}}.

\bibitem[Wang et~al.(2023{\natexlab{c}})Wang, Zhu, Huang, Chen, and Lu]{wang2023drivedreamer}
Xiaofeng Wang, Zheng Zhu, Guan Huang, Xinze Chen, and Jiwen Lu.
\newblock Drivedreamer: Towards real-world-driven world models for autonomous driving.
\newblock \emph{arXiv preprint arXiv:2309.09777}, 2023{\natexlab{c}}.

\bibitem[Wang et~al.(2022)Wang, Vitor~Campagnolo, Zhang, Zhao, and Solomon]{wang2022detr3d}
Yue Wang, Guizilini Vitor~Campagnolo, Tianyuan Zhang, Hang Zhao, and Justin Solomon.
\newblock Detr3d: 3d object detection from multi-view images via 3d-to-2d queries.
\newblock In \emph{In Conference on Robot Learning}, pages 180--191, 2022.

\bibitem[Wu et~al.(2022{\natexlab{a}})Wu, Escontrela, Hafner, Abbeel, and Goldberg]{22DayDreamer}
Philipp Wu, Alejandro Escontrela, Danijar Hafner, Pieter Abbeel, and Ken Goldberg.
\newblock Daydreamer: World models for physical robot learning.
\newblock In \emph{Conference on Robot Learning, CoRL 2022, 14-18 December 2022, Auckland, New Zealand}, 2022{\natexlab{a}}.

\bibitem[Wu et~al.(2022{\natexlab{b}})Wu, Jia, Chen, Yan, Li, and Qiao]{wu2022trajectory}
Penghao Wu, Xiaosong Jia, Li Chen, Junchi Yan, Hongyang Li, and Yu Qiao.
\newblock Trajectory-guided control prediction for end-to-end autonomous driving: A simple yet strong baseline.
\newblock \emph{Advances in Neural Information Processing Systems}, 35:\penalty0 6119--6132, 2022{\natexlab{b}}.

\bibitem[Xie et~al.(2023)Xie, Chen, Hong, and Liu]{xie2023citydreamer}
Haozhe Xie, Zhaoxi Chen, Fangzhou Hong, and Ziwei Liu.
\newblock Citydreamer: Compositional generative model of unbounded 3d cities.
\newblock \emph{arXiv preprint arXiv:2309.00610}, 2023.

\bibitem[Xu et~al.(2023)Xu, Zhang, Xie, Zhao, Guo, Wong, Li, and Zhao]{DriveGPT4}
Zhenhua Xu, Yujia Zhang, Enze Xie, Zhen Zhao, Yong Guo, Kwan{-}Yee.~K. Wong, Zhenguo Li, and Hengshuang Zhao.
\newblock Drivegpt4: Interpretable end-to-end autonomous driving via large language model.
\newblock \emph{CoRR}, abs/2310.01412, 2023.

\bibitem[Yang et~al.(2023)Yang, Ma, Peng, Guo, Lin, and Yu]{yang2023bevcontrol}
Kairui Yang, Enhui Ma, Jibin Peng, Qing Guo, Di Lin, and Kaicheng Yu.
\newblock Bevcontrol: Accurately controlling street-view elements with multi-perspective consistency via bev sketch layout.
\newblock \emph{arXiv preprint arXiv:2308.01661}, 2023.

\bibitem[Zeng et~al.(2022)Zeng, Liu, Du, Wang, Lai, Ding, Yang, Xu, Zheng, Xia, et~al.]{zeng2022glm}
Aohan Zeng, Xiao Liu, Zhengxiao Du, Zihan Wang, Hanyu Lai, Ming Ding, Zhuoyi Yang, Yifan Xu, Wendi Zheng, Xiao Xia, et~al.
\newblock Glm-130b: An open bilingual pre-trained model.
\newblock \emph{arXiv preprint arXiv:2210.02414}, 2022.

\bibitem[Zhang et~al.(2023)Zhang, Li, and Bing]{zhang2023videollama}
Hang Zhang, Xin Li, and Lidong Bing.
\newblock Video-llama: An instruction-tuned audio-visual language model for video understanding.
\newblock \emph{arXiv preprint arXiv:2306.02858}, 2023.

\bibitem[Zhu et~al.(2023)Zhu, Chen, Shen, Li, and Elhoseiny]{zhu2023minigpt}
Deyao Zhu, Jun Chen, Xiaoqian Shen, Xiang Li, and Mohamed Elhoseiny.
\newblock Minigpt-4: Enhancing vision-language understanding with advanced large language models.
\newblock \emph{arXiv preprint arXiv:2304.10592}, 2023.

\end{thebibliography}
}


\end{document}